%% file: main.tex
\documentclass{article}
\usepackage{ijcai22}

\usepackage{times}
\usepackage{soul}
\usepackage{url}
\usepackage[hidelinks]{hyperref}
\usepackage[utf8]{inputenc}
\usepackage[small]{caption}
\usepackage{graphicx}
\usepackage{amsmath}
\usepackage{amsthm}
\usepackage{booktabs}
\usepackage{algorithm}
\usepackage{algorithmic}
\usepackage{float}
\usepackage{multirow}
\usepackage{graphicx}

\title{SideRT: A Real-time Pure Transformer Architecture for Single Image Depth Estimation}

\author{
Chang Shu\and
Ziming Chen\and
Lei Chen\and
Kuan Ma\and
Minghui Wang\And
Haibing Ren
\affiliations
Meituan Group\\
\emails
\{shuchang02,chenziming02,makuan@meituan.com,chenlei90,wangminghui08,renhaibing\}@meituan.com
}
\begin{document}

\maketitle

\input 0abstract
\input 1introduction
\input 2relatedwork

\input 3method
\input 4experiment

\input 5conclusion

\bibliographystyle{named}
\bibliography{reference}
\end{document}

%% file: 0abstract.tex
\begin{abstract}
%our method improves state-of-the-art performance from 0.058 to 0.054 on KITTI and from 0.103 to 0.093 on NYU .
%Comprehensive experiments and detailed analysis via visualization demonstrate the effectiveness of the proposed CSA module.
% Predicting depth from a single image is an ill-posed problem and is essential for 3D geometry understanding.
% CNN-based architectures have substantially pushed forward the state-of-the-art.
% Since context modeling is critical for this task, most methods put tremendous effort into obtaining global context, where large backbones, feature pyramid fusion, dilated convolution are adopted.
% Compared to the locality of convolutions, attention mechanisms or transformers originally designed for capturing long-range dependencies might be a better choice.
% More and more newest attention mechanisms and transformers are introduced to this field.
% However, integrating those components usually complicates architectures and could lead to a decrease in inference speed.

Since context modeling is critical for estimating depth from a single image, researchers put tremendous effort into obtaining global context.
Many global manipulations are designed for traditional CNN-based architectures to overcome the locality of convolutions.
Attention mechanisms or transformers originally designed for capturing long-range dependencies might be a better choice, but usually complicates architectures and could lead to a decrease in inference speed.
In this work, we propose a pure transformer architecture called SideRT that can attain excellent predictions in real-time.
In order to capture better global context, Cross-Scale Attention (CSA) and Multi-Scale Refinement (MSR) modules are designed to work collaboratively to fuse features of different scales efficiently.
CSA modules focus on fusing features of high semantic similarities, while MSR modules aim to fuse features at corresponding positions.
These two modules contain a few learnable parameters without convolutions, based on which a lightweight yet effective model is built. 
This architecture achieves state-of-the-art performances in real-time (51.3 FPS) and becomes much faster with a reasonable performance drop on a smaller backbone Swin-T (83.1 FPS).
Furthermore, its performance surpasses the previous state-of-the-art by a large margin, improving AbsRel metric 6.9\% on KITTI and 9.7\% on NYU.
To the best of our knowledge, this is the first work to show that transformer-based networks can attain state-of-the-art performance in real-time in the single image depth estimation field.
Code will be made available soon.
\end{abstract}

%% file: 1introduction.tex
\section{Introduction}
%单幅图像深度预测的重要性single image depth estimation（SIDE）
%CNN架构一度统治CV领域，包括depth领域
%side领域的特点：缺少几何约束 
%依赖模型强大的学习能力来获取对全局结构的理解
%side领域有大感受野大backbone发展的趋势
%举例说明 引用论文：dorn vip-deeplab DPT
%但是这么干不是最优的，transformer是更好的解决方案。
%对于近些年来，在nlp领域风靡一时transformer架构开始在cv领域展现优越性。例如：
%transformer的好处：参考综述
%transformer的优点适合side任务
%例如：DPT
%但是混合结构CNN+transformer
%在纯transformer领域，还没有人做过探究
%阐明纯transformer的困难
%我们提出：
%创新点\item
%实验表明

Single image depth estimation (SIDE) has a pivotal role in extracting 3D geometry, which has a wide range of practical applications, including automatic driving, robotics navigation, and augmented reality.
The main difficulty of SIDE is that: unlike other 3D vision problems, multiple views are missing to establish the geometric relationship as 3D geometric clues can only be dug from a single image.

\begin{figure}[!t]
\centering
\includegraphics[width=3in]{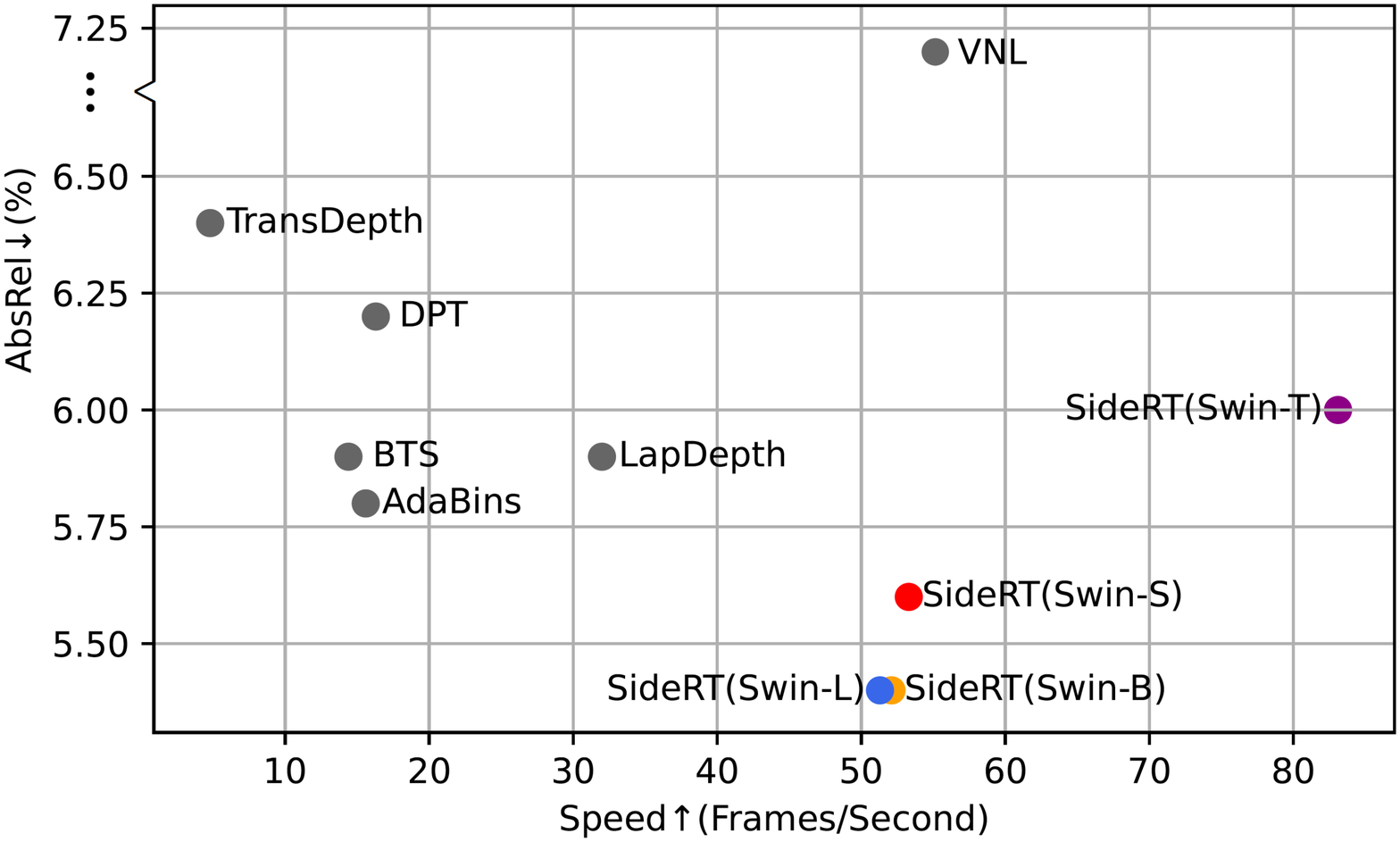}
\caption{
Inference speed versus AbsRel performance on the test set of the KITTI dataset.
Previous state-of-the-art models are marked as grey points. 
Note that SideRT models achieve better accuracy with state-of-the-art methods at a much faster speed.}
\label{accspeed}
\end{figure}

In order to solve this ill-posed problem, the ability to extract global context is paid tremendous attention, which largely relies on the powerful learning capability of modern deep neural networks.
CNN-based architectures \cite{eigen,dorn,bts,vip-deeplab} once dominate the SIDE field, due to the intrinsic locality of convolution, global context is only obtained near the bottleneck.
The global context is usually maintained in low-resolution feature maps. 
Essential clues for 3D structures like local details are lost after consecutive convolutional operations.
To obtain high-resolution global context, there has been a trend in SIDE field to enlarge receptive field via large backbones \cite{densenet,resnext,hrnet}, feature pyramid \cite{fpn}, spatial pyramid pooling \cite{spp} and atrous convolution \cite{aspp,denseaspp}.

Another paradigm to extract global context is taking advantage of the long-range dependency modeling capability of the attention mechanisms.
The attention module \cite{transformer,nonlocal} computes the responses at each position by estimating matching scores to all positions and gathering the corresponding embeddings accordingly, so a global receptive field is guaranteed.
Using attention as the main component, transformers which are initially designed for natural language processing, are found more and more applications in the computer vision field \cite{vit,swin,detr}.
Thanks to the powerful ability to establish long-range dependencies of attention mechanisms and transformers, integrating them into fully convolutional architectures \cite{dpt,adabins,transdepth} has pushed state-of-the-art performance forward a lot.

Since the attention mechanism is usually time- and memory-consuming, inference speed has to be compromised when using transformers or attention mechanisms.
Many works have been devised for more efficient implementation, but similar works are rare in the SIDE field.

This paper explores how to achieve state-of-the-art performance in real-time when using transformers and attention mechanisms.
We introduce the SIDE Real-time Transformer (SideRT) based on an encoder-decoder architecture.
Swin transformers are used as the encoder. 
The decoder is built on a novel attention mechanism named Cross-Scale Attention (CSA) and a Multi-Scale Refinement module (MSR).
Both CSA and MSR modules are global operations and work collaboratively.
In CSA modules, finer-resolution features are augmented by coarser-resolution features according to attention scores defined by semantic similarity.
In MSR modules, coarser-resolution features are merged to spatially corresponding finer-resolution features.
Since a few learnable parameters are used in the proposed modules, feature maps at different scales are fused with a fair computational overhead.
Based on CSA and MSR modules, we build a lightweight decoder that conducts hierarchical depth optimization progressively to get the final prediction in a coarse-to-fine manner.
Furthermore, Multi-Stage Supervision (MSS) is added at each stage to ease the training process.

As depicted in Figure \ref{accspeed}, the proposed SideRT significantly outperforms the previous state-of-the-art at a speed of 51.3 FPS. 
It improves the AbsRel metric from 0.058 to 0.054 on KITTI and from 0.103 to 0.093 on NYU.
Moreover, SideRT can achieve 0.060 AbsRel on KITTI, and 0.124 AbsRel on NYU on smaller backbone Swin-T \cite{swin} at a speed of 83.1 FPS and 84.4 FPS respectively.
To the best of our knowledge, this is the first work to show that transformer-based networks can attain state-of-the-art performance in real-time in the single image depth estimation field.

% We borrow fine ideas from building fully convolutional networks for transformer-based architectures design.
% To compensate the drawback of attention mechanism that it lacks of spatial inductive bias in capturing local information, we augment the feature coming after CSA by adding boundaries details computed from laplacian pyramid of input images.
% To ease the training process, we add hierarchical supervision at each stage and design a residual prediction mechanism to progressively get final prediction in a coarse-to-fine way.

% Our contributions can be summarized as below:
% \begin{itemize}
% \item We propose a novel attention mechanism for multi-scale feature fusion, which add only a little computational overall, based on which, we build a real-time transformer-based architecture for SIDE, setting a new state-of-the-art performance on two challenging benchmarks, KITTI~\cite{kitti} and NYU~\cite{nyu}.
% \item We empirically verify several techniques originally designed for fully convolutional networks, and integrate them to improve the performance of the transformer-based architectures.
% \item 
% \end{itemize}

%% file: 2relatedwork.tex
% !TEX root = main.tex
\section{Related Work}
\label{rw}

SIDE is a practical vision task for 3D scene understanding. 
Due to the ambiguity of 3D mapping in a single view, SIDE is ill-posed. 
However, with the help of deep learning, considerable progress has been made in the SIDE field.

% \subsection{CNN-based}
% \cite{eigen} firstly brings the convolution neural network (CNN)~\cite{alexnet} to the SIDE task, and subsequent researchers apply the fully convolutional network in more powerful network architectures, like ResNet~\cite{resnet} and DenseNet~\cite{densenet}, for depth estimation. The original intention is to allow the network to learn better the global scene information, which helps the model understand the depth distribution of the image scene and draw a more reliable inference. The multi-scale approach, represented by the encoder-decoder architecture, gradually reduces the resolution, extracts global contextual knowledge from low resolution, and then uses skip-connection to help restore high-frequency information~\cite{big2small}. \cite{big2small,dorn,song2021monocular} use atrous spatial pyramid pooling module (ASPP) to extract features, which can fuse multi-level image information without loss of resolution. Some studies have also tried to optimize depth map by global information propagation strategies, such as CRF~\cite{li2015depth} and CSPN~\cite{cspn}. In addition to network structure improvement, incorporating scene information is also beneficial for depth inference. \cite{sdc-depth} takes advantage of semantic information and instance information to decompose inference depth and simplify prediction. \cite{song2021monocular} introduces the Laplacian pyramid to extract the object boundary, as well as global layout, to assist depth estimation.
\textbf{CNN-based.} \cite{eigen} firstly brings CNNs to the SIDE task, and subsequent researchers introduce more powerful networks, like ResNet and DenseNet, for depth estimation. 
They aim to encourage networks to learn the global context better, which helps the model understand the depth distribution of the image scene and draw a more reliable inference. 
Many global manipulations like atrous spatial pyramid pooling module (ASPP), spatial pyramid pooling and feature pyramid are adopted to enlarge receptive field \cite{bts,dorn,song2021monocular}.

\textbf{Attention-based.} The attention mechanism can establish associations between all the pixels, thereby overcoming the problem of establishing long-range dependencies. 
This capability is favored by SIDE researchers for contextual knowledge extraction. 
\cite{pwa} takes advantage of the attention mechanism to perform feature learning on the patches to obtain higher prediction accuracy. 
\cite{xu2021pyramid} designs channel attention and spatial attention modules to further improve the high-level context features and low-level spatial features. 
\cite{huynh2020guiding} explores the self-attention mechanism to establish pixel's association and treats this information as depth prior.
\cite{bidirectional,hao2018detail} apply the attention mechanism to fuse multi-level features. \cite{jiao2018look} focuses on the distribution of depth prediction data and designs an attention-driven loss to improve the quality of depth prediction in long range.

% \subsection{transformer-based}

% In the NLP, the transformer mechanism can obtain the correlation between all vectors and highlight critical information, which makes it have wide application ~\cite{transformer,dai2019transformer}. Vision-transformer (ViT) has been popular in computer vision for its excellent global image processing capability in the past two years. It promotes network learning highly representative features and benefits the task understanding of computer vision. In addition to achieving excellent results in various visual tasks~\cite{swin,hu2019local,carion2020end,dosovitskiy2020image}，ViT also performs brilliantly on the SIDE. \cite{yang2021transformer} proposes the ViT architecture for SIDE to develop the feature learning capability of transformer. In order to capture local-level details, they also design an attention module,  which consists of channel attention and spatial attention. DPT~\cite{dpt} leverages the ViT to replace the convolution layers to overcome its limited receptive field on feature learning and prove the ViT-based encoder is effective than the convolution-based in dense prediction tasks. The aim of AdaBins~\cite{adabins} is to implement adaptive splitting of the depth range, and ViT can learn better splitting through its global information processing on high-resolution features. This strategy also yields excellent results.

\textbf{Transformer-based.} 
Transformer \cite{transformer}, first applied to the NLP field, is a type of deep neural network mainly based on the self-attention mechanism. 
Thanks to its strong representation capabilities, researchers are looking for ways to apply transformer in the SIDE tasks.
\cite{transdepth} introduces ViT architectures \cite{vit} to this field to compensate the intrinsic locality of convolutions. 
\cite{dpt} leverages ViT in place of convolutional networks as a backbone and prove the ViT-based encoder is more effective than the convolution-based for SIDE. 
\cite{adabins} utilizes ViT to perform global processing of the scene’s information and subsequently learn adaptive dividing of the depth range. 

\begin{figure*}[!t]
\centering
\includegraphics[width=\linewidth]{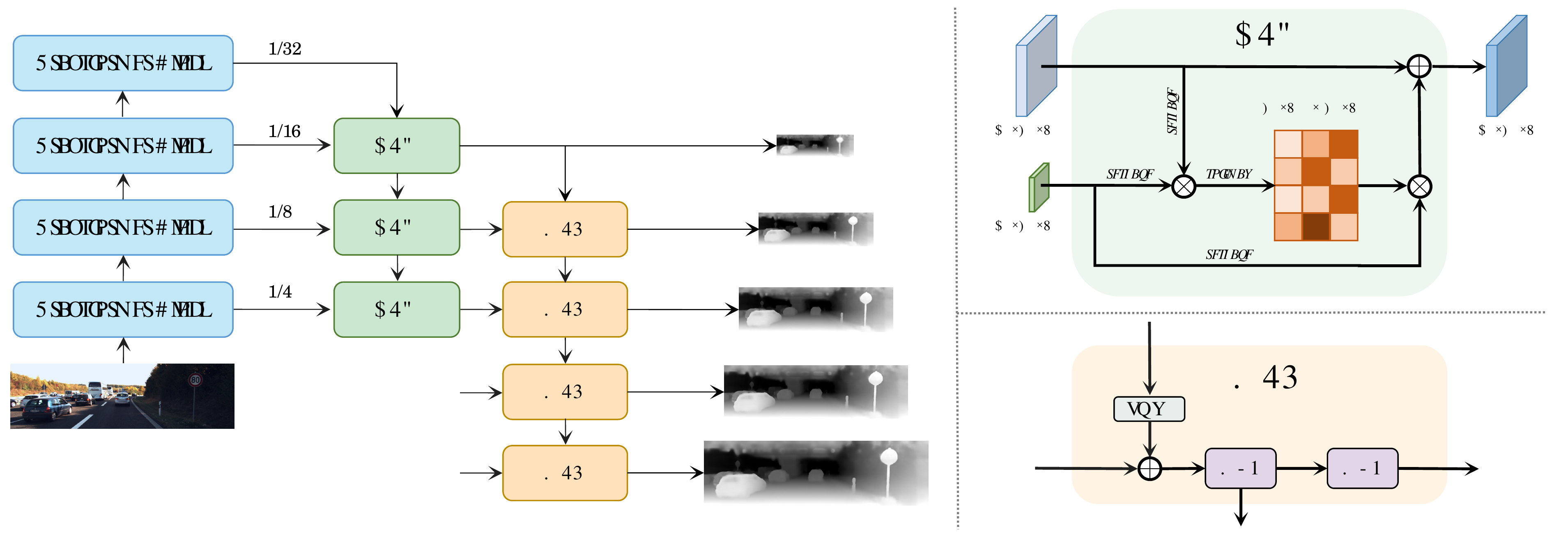}
\caption{(1) Architecture overview. The input image is sent into a series of transformer blocks to obtain a feature pyramid. 
CSA and MSR modules progressively merge feature maps from adjacent scales to obtain more powerful feature representations. `0' means no input.
(2) In CSA modules, finer-resolution features are augmented by coarser-resolution features with matching scores defined by semantic similarity. 
$\bigoplus$ and $\bigotimes$ respectively denote element-wise sum and matrix multiplication.
(3) In MSR modules, coarser-resolution features are merged to spatially corresponding finer-resolution features.}
\label{overall}
\end{figure*}

Unlike prior works which use either fully convolutional networks or combine CNNs with transformers and attention mechanisms, we explore the possibility of building the architecture without convolutions.
Furthermore, we prove that the state-of-the-art can still be achieved even without convolutions.
We also demonstrate that even using heavy backbones like Swin-L, our model can still run in real-time.
% architecture propose a new SIDE architecture which is a pure transformer network. 
% To be specific, we utilize swin transformers \cite{swin} to extract powerful representative features firstly, which are rich in global information of image scene. Then, cross-scale attention (CSA) is proposed to aggregate the diverse information from multi-level features. Accompanied by the multi-scale refinement and multi-stage supervision, the transformer and CSA module do better in feature learning and information fusion. The proposed network is convolution-free yet owning excellent efficiency and effectiveness, which can support real-time prediction, as well as ranking first on the classic dataset (i.e., KITTI and NYU).

%% file: 3method.tex
\section{Method}
\label{method}

In this section, we first present a general framework of our network and then introduce in detail three major components of our networks, which are cross-scale attention, multi-scale refinement and multi-stage supervision.
Figure \ref{overall} illustrates our SideRT architecture.

\subsection{Overview}

Our proposed SideRT has a simple yet efficient encoder-decoder architecture that predicts depth from a single image. 
We adopt Swin Transformers \cite{swin} as our backbones, each image is divided into several $4 \times 4$ non-overlapping patches.
The patch, along with relative positional embedding, is embedded by the linear projection layer and the result will be processed by a series of shift window based transformer blocks. 
Feature maps from all four stages will be utilized during decoding.

The input to the decoder is a set of multi-scale feature maps of four stages: (I) stage 1, $\frac{H}{4} \times \frac{W}{4} \times C $, 
(II) stage 2, $\frac{H}{8} \times \frac{W}{8} \times 2C $, 
(III) stage 3, $\frac{H}{16} \times \frac{W}{16} \times 4C $, 
and (IV) stage 4, $\frac{H}{32} \times \frac{W}{32} \times 8C $.
$H$ and $W$ respectively represent the height and width of input images, $C$ denotes the number of channels of features from stage 1.
Analogous with a standard feature pyramid, the decoder fuses feature maps progressively in a coarse-to-fine way. 
Our decoder consists two basic modules: cross-scale attention modules and multi-scale refinement modules (i.e. CSA and MSR modules in Figure \ref{overall}). 
To obtain global context, CSA modules aim to fuse feature maps following the guidance of semantic similarity, while MSR modules aim to fuse feature maps according to spatial corresponding relationship.
Fusion operations are conducted in a coarse-to-fine manner to get the final prediction, which keeps the same resolution as the input images.  

\begin{table*}[!t]
\centering
\resizebox{\textwidth}{!}{
\begin{tabular}{@{}llcccccccrr@{}}
\toprule
\multirow{2}{*}{Method} & \multirow{2}{*}{Backbone} & \multirow{2}{*}{AbsRel↓} & \multirow{2}{*}{SqRel↓} & \multirow{2}{*}{RMSE↓} & \multirow{2}{*}{RMSE log↓} & \multirow{2}{*}{$\delta_1$↑} & \multirow{2}{*}{$\delta_2$↑} & \multirow{2}{*}{$\delta_3$↑} & \multirow{2}{*}{Params↓} & \multirow{2}{*}{FPS↑} \\
                        &                           &                          &                         &                        &                            &                              &                              &                              &                          &                       \\ \midrule
VNL                     & ResNext-101               & 0.072                    & -                       & 3.258                  & 0.117                      & 0.938                        & 0.990                        & 0.998                        & 90.4 M                   & 54.8                  \\
DORN                    & ResNet-101                & 0.072                    & 0.307                   & 2.727                  & 0.120                      & 0.932                        & 0.984                        & 0.994                        & -                        & -                     \\
TransDepth              & ResNet-50+ViT             & 0.064                    & 0.252                   & 2.755                  & 0.098                      & 0.956                        & 0.994                        & \textbf{0.999}               & 247.4 M                  & 4.8                   \\
DPT                     & VIT-Hybrid                & 0.062                    & -                       & 2.573                  & 0.092                      & 0.959                        & 0.995                        & \textbf{0.999}               & 123.0 M                  & 16.4                  \\
BTS                     & ResNext-101               & 0.059                    & 0.245                   & 2.756                  & 0.096                      & 0.956                        & 0.993                        & 0.998                        & 112.8 M                  & 14.4                  \\
LapDepth                & ResNext-101               & 0.059                    & 0.212                   & 2.446                  & 0.091                      & 0.962                        & 0.994                        & \textbf{0.999}               & 73.0 M                   & 32.0                  \\
AdaBins                 & EfficientNet-B5           & 0.058                    & 0.190                   & 2.360                  & 0.088                      & 0.964                        & 0.995                        & \textbf{0.999}               & 78.0 M                   & 15.6                  \\ \midrule
\multirow{4}{*}{Ours}   & Swin-T                    & 0.060                    & 0.206                   & 2.441                  & 0.093                      & 0.959                        & 0.995                        & \textbf{0.999}               & \textbf{28.6 M}          & \textbf{83.1}         \\
                        & Swin-S                    & 0.056                    & 0.187                   & 2.306                  & 0.087                      & 0.965                        & 0.996                        & \textbf{0.999}               & 50.1 M                   & 53.3                  \\
                        & Swin-B                    & \textbf{0.054}           & \textbf{0.170}          & \textbf{2.212}         & 0.083                      & 0.971                        & 0.996                        & \textbf{0.999}               & 89.2 M                   & 52.1                  \\
                        & Swin-L                    & \textbf{0.054}           & 0.173                   & 2.249                  & \textbf{0.082}             & \textbf{0.972}               & \textbf{0.997}               & \textbf{0.999}               & 200.4 M                  & 51.3                  \\ \bottomrule
\end{tabular}
}
\caption{
Comparison to the state-of-the-art on KITTI dataset, best results are in bold.
}
\label{kittisota}
\end{table*}

\begin{table*}[!t]
\centering
\resizebox{\textwidth}{!}{
\begin{tabular}{@{}llccccccrr@{}}
\toprule
\multirow{2}{*}{Method} & \multirow{2}{*}{Backbone} & \multirow{2}{*}{AbsRel↓} & \multirow{2}{*}{RMSE↓} & \multirow{2}{*}{log10↓} & \multirow{2}{*}{$\delta_1$↑} & \multirow{2}{*}{$\delta_2$↑} & \multirow{2}{*}{$\delta_3$↑} & \multirow{2}{*}{Params↓} & \multirow{2}{*}{FPS↑} \\
                        &                           &                          &                        &                         &                              &                              &                              &                          &                       \\ \midrule
SharpNet                & ResNet-50                 & 0.139                    & 0.495                  & 0.047                   & 0.888                        & 0.979                        & 0.995                        & 114.1 M                  & 156.7                 \\
DORN                    & ResNet-101                & 0.115                    & 0.509                  & 0.051                   & 0.828                        & 0.965                        & 0.992                        & -                        & -                     \\
LapDepth                & ResNext-101               & 0.110                    & 0.393                  & 0.047                   & 0.885                        & 0.979                        & 0.995                        & 73.0 M                   & 40.3                  \\
BTS                     & DenseNet-161              & 0.110                    & 0.392                  & 0.047                   & 0.885                        & 0.978                        & 0.994                        & 47.0 M                   & 24.5                  \\
DPT                     & VIT-Hybrid                & 0.110                    & 0.357                  & 0.045                   & 0.904                        & 0.988                        & 0.998                        & 123.0 M                  & 24.3                  \\
VNL                     & ResNext-101               & 0.108                    & 0.416                  & 0.048                   & 0.875                        & 0.976                        & 0.994                        & 90.4 M                   & 53.6                  \\
TransDepth              & ResNet-50+ViT             & 0.106                    & 0.365                  & 0.045                   & 0.900                        & 0.983                        & 0.996                        & 247.4 M                  & 6.5                   \\
AdaBins                 & EfficientNet-B5           & 0.103                    & 0.364                  & 0.044                   & 0.903                        & 0.984                        & \textbf{0.997}               & 78.0 M                   & 19.9                  \\ \midrule
\multirow{4}{*}{Ours}   & Swin-T                    & 0.124                    & 0.428                  & 0.052                   & 0.860                        & 0.974                        & 0.994                        & \textbf{28.6 M}          & \textbf{84.4}         \\
                        & Swin-S                    & 0.108                    & 0.380                  & 0.046                   & 0.892                        & 0.982                        & 0.996                        & 50.1 M                   & 53.2                  \\
                        & Swin-B                    & 0.100                    & 0.354                  & 0.043                   & 0.908                        & 0.985                        & \textbf{0.997}               & 89.2 M                   & 51.1                  \\
                        & Swin-L                    & \textbf{0.093}           & \textbf{0.335}         & \textbf{0.040}          & \textbf{0.922}               & \textbf{0.990}               & \textbf{0.997}               & 200.4 M                  & 50.3                  \\ \bottomrule
\end{tabular}
}
\caption{
Comparison to the state-of-the-art on NYU dataset, best results are in bold.
}
\label{nyusota}
\end{table*}

\subsection{Cross-Scale Attention}

Our proposed CSA module consists of two parts: a linear layer to project each feature to the same number of channels and an attention-based fusion to fuse feature maps from adjacent scales according to semantic similarity.
Denote input feature maps as $F_1 \in C_1 \times H_1 \times W_1$ and $F_2 \in C_2 \times H_2 \times W_2$, while $F_1$ is from the shallower stage and $F_2$ is from the deeper stage of the encoder.  
After linear projection, we propose an attention-based method to fuse these two feature maps:
\begin{align}
F_{12} & = L(F_1) + softmax(L(F_1)\times L(F_2)) \times L(F_2)
\end{align}
where $L(\cdot)$ is the linear projection operation. 
The linear layer will project $F_1$ from $C_1 \times H_1 \times W_1$ to $C \times (H_1\!\times\!W_1)$, and $F_2$ from $C_2 \times H_2 \times W_2$ to $ C \times (H_2\!\times\!W_2)$ respectively.

For traditional CNN-based methods, global context information only exists near the encoder bottleneck and will be gradually weakened during the hierarchical upsampling of the decoder.
Different from them, this CSA module is able to naturally capture global context dependency between feature maps from adjacent scales through calculating attention scores $softmax(L(F_1)\times L(F_2))$ across the entire feature map. 
Furthermore, this global context dependency contains both semantic-related and depth-related information, as illustrated in the visual analysis of Section \ref{visana}.

\subsection{Multi-Scale Refinement}

Several multi-scale refinement (MSR) modules work collaboratively in a top-down hierarchy pyramid pattern. 
A coarser-resolution feature map from the higher MSR pyramid level and a finer-resolution feature map from the lower CSA pyramid level are fed into a MSR module.
The MSR module outputs a refined feature map with the help of a low-level semantic but more accurately-localized feature map from the CSA pyramid. 
The output feature map of the MSR module and the next finer-resolution feature map from the lower CSA pyramid level will be fed into the next MSR module until the final feature map is generated. 
In a word, the whole refinement process is iterated in a coarse-to-fine way.

Each multi-scale refinement module includes three parts: a bilinear interpolation to upsample coarser-resolution feature map from MSR pyramid, an element-wise addition between the upsampled feature map and finer-resolution feature map from the CSA pyramid, and two MLP layers to reduce aliasing effect of upsampling and generate a depth map at each scale.
The last two MSR modules are appended to generate the final result with the input resolution.
In particular, unlike other MSR modules, they do not receive outputs from CSA modules.

Different from the fusion style of CSA modules which fuse features according to semantic similarity, MSR modules leverage upsampling to fuse features with corresponding spatial positions.
Furthermore, the introduction of MLP layers facilitates information to flow globally. 

\subsection{Multi-Stage Supervision}

In order to ease the training of the early stages of the whole architecture, we propose a multi-stage supervision (MSS) strategy at each scale to supervise the training process. 
In each stage, the loss between the prediction and the corresponding ground truth is calculated. 
Finally, multi-stage losses are weighted summed together.

Due to the limitation of 3D sensors, the depth data is dense at close areas whereas very sparse in the distance.
To alleviate this imbalance problem, we adopt the square root loss function introduced in \cite{song2021monocular}.
This loss calculates the difference of predicted depth value and the ground truth in the log space, as shown below:
\begin{align}
    L\left(y, y^{*}\right) &=\sqrt{\frac{1}{n} \sum_{i \in V}^{N_{V}} d_{i}^{2}-\frac{\lambda}{n^{2}}\left(\sum_{i \in V}^{N_{V}} d_{i}\right)^{2}} \\
    d_{i} &=\log y_{i}-\log y_{i}^{*} 
\end{align}
where $y$ and $y^{*}$ respectively represent the predicted depth map and the ground truth, $V$ is the set of valid pixels in the depth map, $N_{V}$ indicates the total number of valid pixels, and the balance coefficient lambda is set to 0.85.

%% file: 4experiment.tex
\begin{figure*}[!t]
\centering
\includegraphics[width=7.2in]{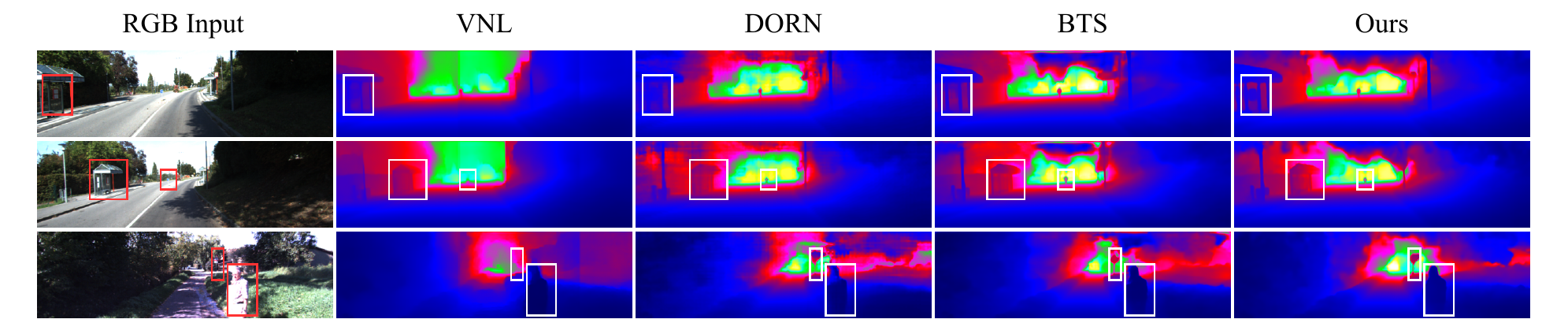}
\caption{Visualization of depth predictions in the KITTI dataset.}
\label{kittivis}
\end{figure*}
\section{Experiments}
\label{exp}

Firstly, we apply SideRT on two challenging datasets: KITTI and NYU. 
For both datasets, we show that SideRT can significantly outperform previous state-of-the-art methods at a much faster speed. 
At the end of this section, we conduct comprehensive ablations of different components and detailed visualization to verify the effectiveness of our method.

\subsection{Implementation Details} 
The model is implemented with PyTorch. 
Swin transformers pretrained on ImageNet \cite{imagenet} are used as encoders. 
The proposed model is trained from scratch for 160 epochs with a batch size of 6 through the AdamW optimizer. 
The weight decaying factor is set to $5e^{-4}$  and the learning rate is set to $1e^{-4}$. 
It takes 8 NVIDIA A100 SXM GPUs in the training process. 
Data augmentation is performed during the training phase to avoid overfitting problems. 
Images from KITTI and NYU are randomly cropped to 704×352 pixels and 512×416 pixels respectively. 
In addition, we randomly adjust the scale factor, rotate the input color images within a specific range and flip input images horizontally with a probability of 0.5. 
The speed of all the methods is tested on a single NVIDIA GeForce RTX 2080 Ti GPU.

\begin{table}[!tp]
\renewcommand\arraystretch{2}
\begin{center}
\begin{tabular}{l}
% \hline
\toprule
$\textbf{AbsRel}:\frac{1}{|D|}\sum_{d \in D}|d^*-d|/d^*$\\ 
$\textbf{RMSE}:\sqrt{\frac{1}{|D|}\sum_{d \in D}||d^*-d||^2}$\\
$\textbf{SqRel}:\frac{1}{|D|}\sum_{d \in D}||d^*-d||^2/d^*$\\
$\textbf{RMSE log}:\sqrt{\frac{1}{|D|}\sum_{d \in D}||logd^*-logd||^2}$\\

$\mathbf{\delta}_\mathbf{t}:\frac{1}{|D|}|\{d \in D| \: max(\frac{d^*}{d},\frac{d}{d^*}) \: < 1.25^t\}|\times 100\%$
\\
\bottomrule
\end{tabular}
\caption{Performance metrics for depth evaluation.
$d$ and $d^*$ respectively denote predicted and ground truth depth, $D$ presents a set of all the predicted depth values of an image, $|.|$ returns the number of the elements in the input set.
}
\label{metric}
\end{center}
\end{table}

\subsection{Benchmark Datasets} 
Two popular datasets (KITTI and NYU) are used for performance evaluation.
The KITTI \cite{kitti} dataset contains the road environment acquired in the autonomous driving scene. 
The resolution of the acquired images is 1242×375 pixels. 
We adopt the split strategy introduced by \cite{eigen} for performance comparison.
The test set contains 697 images from 29 scenes, and the training set contains 23488 images from 32 scenes. 
The maximum value of prediction depth is 80 meters.

The NYU dataset \cite{nyu} contains 120K images obtained by Microsoft Kinect camera, including 464 indoor scenes with a resolution of 640×480. 
We also follow \cite{eigen} to set train/test split, which includes 249 scenarios for training and 654 images from remaining 215 scenarios for testing. 
The depth map is center-cropped into 561×427 when evaluating the performance.

\begin{figure}[!t]
\centering
\includegraphics[width=3.4in]{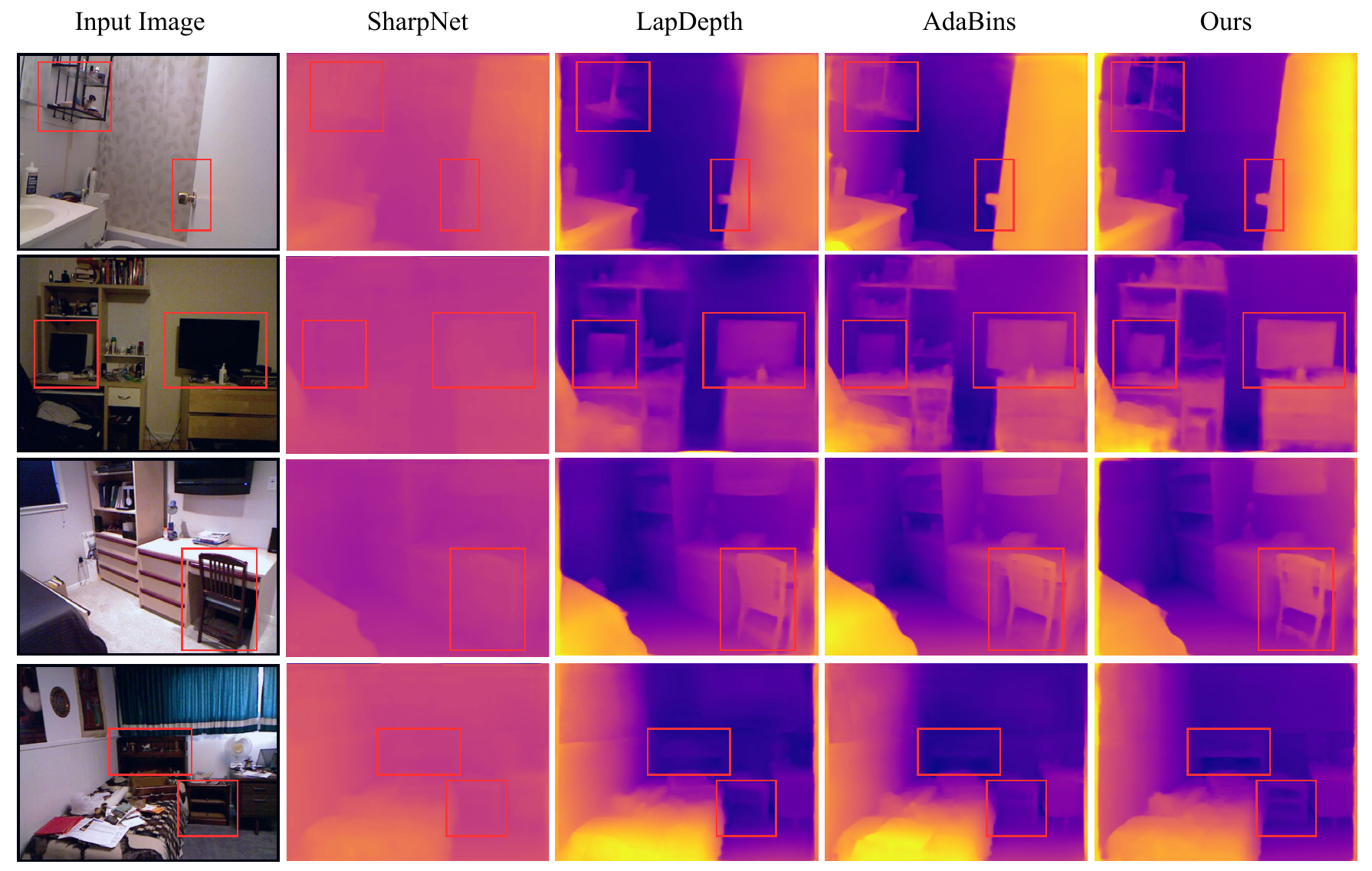}
\caption{Visualization of depth predictions on the NYU dataset}
\label{nyuvis}
\end{figure}

\subsection{Comparison with State-of-the-art} 
\label{sota}
In order to quantify the performance of our model, we used metrics provided by \cite{eigen}, which are widely applied in the performance evaluation of monocular depth estimation.
The specific formulations of these metrics are shown in table \ref{metric}.

We compared our method with state-of-the-art methods on the KITTI and NYU datasets on these metrics. 
As shown in Table \ref{kittisota} and \ref{nyusota}, it is obvious that our method achieves the best performance on all metrics.
On the KITTI dataset, compared with the previous state-of-the-art, AbsRel has decreased by 6.9\% and SqRel has decreased by 8.9\%. 
On the NYU dataset, compared with the previous state-of-the-art, AbsRel decreased by 9.7\% and RMSE decreased by 8.0\%. 

\begin{table*}[!t]
\centering
\resizebox{\textwidth}{!}{
\begin{tabular}{@{}lcccccccrr@{}}
\toprule
Method             & AbsRel ↓ & SqRel ↓ & RMSE ↓ & RMSE log ↓ & $\delta_1$↑ & $\delta_2$↑ & $\delta_3$↑ & Params↓ & FPS ↑ \\ \midrule
Swin-T             & 0.239    & 1.786   & 6.304  & 0.301      & 0.638       & 0.864       & 0.950       & 27.5 M  & 104.4 \\
Swin-T+CSA         & 0.077    & 0.309   & 2.969  & 0.117      & 0.932       & 0.988       & 0.997       & 28.3 M  & 92.4  \\
Swin-T+CSA+MSS     & 0.073    & 0.283   & 2.881  & 0.112      & 0.937       & 0.991       & 0.998       & 28.4 M  & 92.4  \\
Swin-T+MSR+MSS     & 0.064    & 0.216   & 2.504  & 0.097      & 0.956       & 0.994       & 0.999       & 27.8 M  & 90.0  \\
Swin-T+CSA+MSR+MSS & 0.060    & 0.206   & 2.441  & 0.093      & 0.959       & 0.995       & 0.999       & 28.6 M  & 83.1  \\ \bottomrule
\end{tabular}
}
\caption{
Ablation study on different components of our work.
}
\label{abla}
\end{table*}

Our small models like SideRT(Swin-T) and SideRT(Swin-S) outperform many state-of-the-art methods in these two challenging benchmarks.
Moreover, we find that the usage of heavy backbone like Swin-L with about 200 million parameters do not slow down the inference process and shows similar inference speed with smaller models like SideRT(Swin-B) and SideRT(Swin-S).

It is worth noting that while surpassing state-of-the-art, we also achieved real-time prediction with a speed of 51.3 FPS. 
For a better understanding of our performance, we visualize predicted depth maps respectively in KITTI and NYU datasets, as shown in Figure \ref{kittivis} and \ref{nyuvis}. 
It can be seen that our method can also successfully predict finely detailed object boundaries that other methods cannot predict clearly.

\subsection{Ablation Study}
To get a better understanding of the contribution of proposed components to the overall performance, an ablation study is performed in Table \ref{abla}.
All the experiments are done on the KITTI dataset and use Swin-T as backbone.
The training and testing strategies are kept the same with Section \ref{sota}.

\textbf{Cross-scale Attention.}
To evaluate the importance of CSA modules, we remove them from the overall architecture (Row 4 and 5 in Table \ref{abla}).
It can be observed that the CSA module improves the performance with a small computational overhead (+0.8M Params) and a small decrease in speed (-12.0 FPS).
Row 1-2 in Table \ref{abla} show that the CSA module significantly improves the depth prediction directly from the encoder Swin-T.
In further visual analysis, we find that the receptive field of the encoder is relatively small, and proposed CSA module will enlarge it tremendously.
It is worth mentioning that comparing performances of Row 2 in Table \ref{abla} and Row 1-2 in Table \ref{kittisota}, the simplest architecture (Swin-T+CSA) gets similar performance with VNL \cite{vnl} and outperforms some metrics of DORN \cite{dorn}.

\textbf{Multi-stage Supervision.}
As mentioned in previous works \cite{mlpmixer,vit,swin}, the training of a pure transformer architecture is much more difficult than a CNN-based counterpart.
After using multi-stage supervision (MSS), we observe that the training loss curve becomes smoother and the model converges more easily.
MSS brings a substantial improvement, including all the metrics, showing that this scheme is very appropriate for accelerating training and promoting the performance of pure transformer architectures.

\textbf{Multi-scale Refinement.}
As shown in Row 3-5 in Table \ref{abla}, introducing MSR modules clearly gives a boost to the performance.
Theoretically speaking, CSA and MSR modules augment original feature maps from encoders in a collaborative way.
CSA focuses on merging features with high similarity from a global prospect.
MSR aims to fuse features with similar positions at different pyramid levels.

\textbf{Inference Speed.}
Table \ref{abla} shows that most of the parameters come from the backbone since our lightweight decoder only contains 1.1 million parameters.
After adding our proposed decoder, the AbsRel metric decreases by 74.9\%, with the inference speed only decreasing by 20.4\%.

\begin{figure}[!t]
\centering
\includegraphics[width=3.4in]{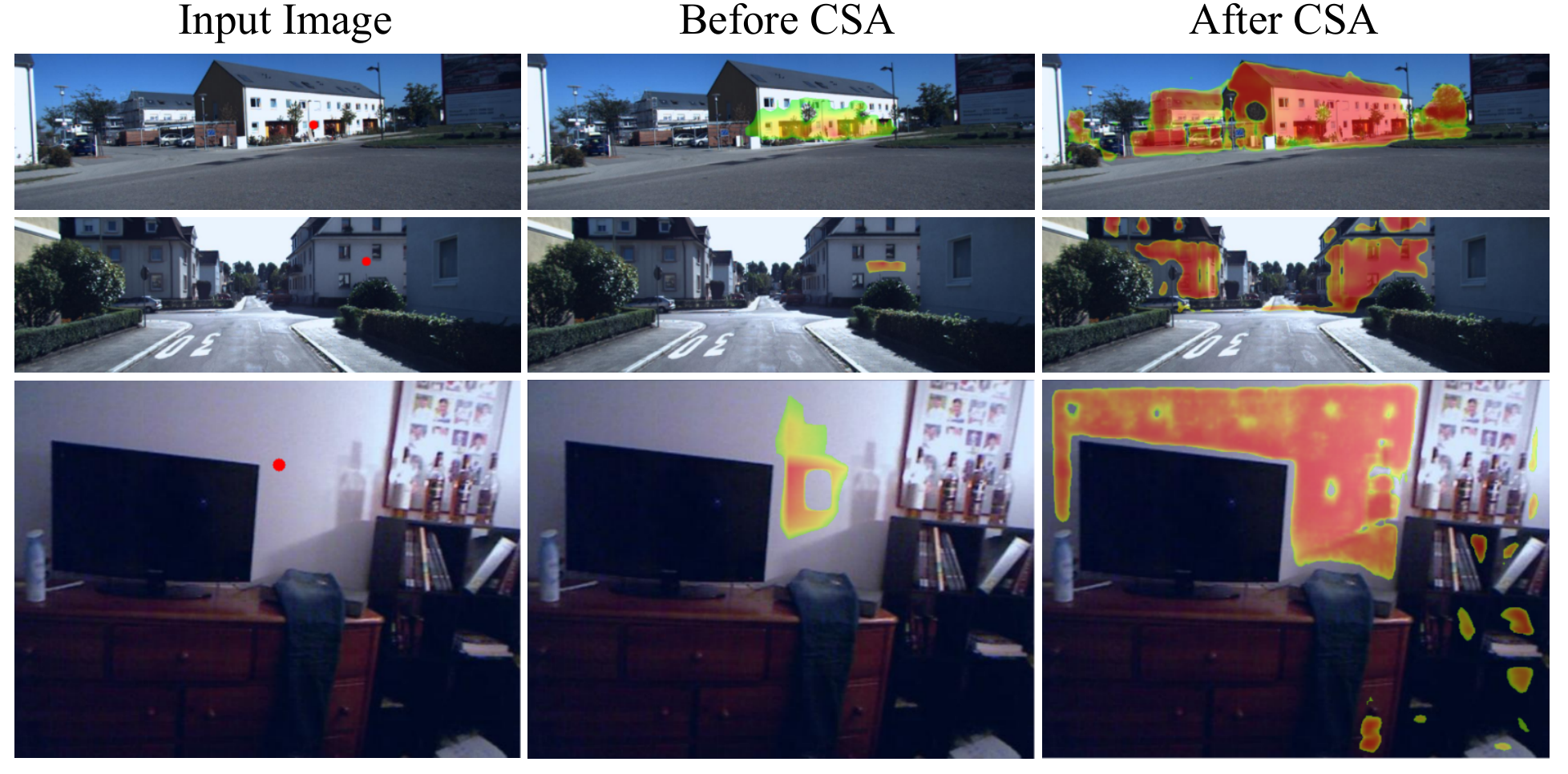}
\caption{Visualization of the receptive field before and after adding CSA modules.
The red dot means the reference pixel. 
Hotter colors indicate stronger correlations.
}
\label{vis}
\end{figure}

\textbf{Visual Analysis.}
\label{visana}
In order to see whether our proposed CSA module actually enlarges the receptive field of the backbone, we follow the common practices \cite{denseaspp,danet} to visualize the empirical receptive field size of a CSA module, as shown in Figure \ref{vis}. 
For an input image, we select a reference pixel (denoted by red dot) and compute its feature similarities with all the other positions.

It is obvious to see that after adding CSA, reference pixels get a stronger response from a larger scope.
We find that CSA modules lead to higher responses from pixels which not only belong to similar categories but also share similar depths.
This is a very beneficial attribute for depth prediction tasks.

% We show the mask located at the red dot for each input picture. 
% We show the mask of two kinds for each point before adding CSA and after adding CSA. 
% We found that after adding CSA, the area of the attention highlight areas gets more extensive, and the value increases, which is reasonable since CSA aggregates features of different dimensions, from local features to global features. Hence, the relevance of points with nearby areas becomes more substantial than the without CSA version.
% We found that our attention mask effectively focused on the associated areas and achieved better performance. For example, in the first row, the red dot mask, which is located on the house, only aggregated the features of part of the house before adding CSA. For windows and doors, the features were not successfully extracted. However, after adding CSA, this point not only aggregates the house's features, including windows and doors, but the mask also covers other houses of similar depth. Other pictures have similar trends.
% The visual mask confirms the effectiveness of our module, in which each position obtains informative contextual information from either the adjacent area or the farther area, which can better extract the features of the adjacent depth area.

%% file: 5conclusion.tex
\section{Conclusion}
\label{con}
This paper proposes two simple yet efficient mechanisms to obtain better global context. 
Based on those techniques, we build a lightweight pure transformer architecture that contains a few learnable parameters without convolutions.
Our model shows powerful context modeling capability, leading to state-of-the-art performances on two challenging datasets.
This work demonstrates that a pure transformer architecture is able to achieve a good trade-off between accuracy and running time efficiency.
These findings will provide insights for future research, encouraging researchers to pay more attention to developing real-time transformer architectures for practical applications.